\documentclass[11pt,a4paper]{article}
\usepackage[]{emnlp2021}

\usepackage{times}
\usepackage{latexsym}

\usepackage{microtype}
\usepackage{times}
\usepackage{latexsym}

\usepackage{soul}
\usepackage{url}
\usepackage[utf8]{inputenc}
\usepackage{graphicx}
\usepackage{amsmath}
\usepackage{booktabs}
\usepackage{colortbl}
\usepackage{multirow}
\usepackage{stfloats}  
\usepackage{setspace}  

\definecolor{ForestGreen}{rgb}{0.13, 0.55, 0.13}

\newcommand{\eat}[1]{}
\newcommand{\red}[1]{\textcolor{red}{#1}}

\newcommand{\todo}[1]{\textcolor{red}{[\textsc{TODO: }#1 ]}}
\newcommand{\camera}[1]{\textcolor{black}{#1}}

\usepackage{quoting}
\newenvironment{myquote}{                   
  \parskip 0mm \begin{quoting}[vskip=0mm,leftmargin=2mm]}{
\end{quoting}}
\newenvironment{ite}{                     
     \parskip 0cm \begin{itemize} \parskip 0cm \parsep 0cm \itemsep 0cm \topsep 0cm}{
        \end{itemize}} 
\newenvironment{enu}{                   
     \parskip 0cm \begin{list}{}{\parsep 0cm \itemsep 0cm \topsep 0cm}}{
       \end{list}} 
\newenvironment{des}{                 
     \parskip 0cm \begin{list}{}{\parsep 0cm \itemsep 0cm \topsep 0cm}}{
       \end{list}} 
\newenvironment{myitemize}{                     
     \parskip 0cm \begin{itemize}{\parsep 0cm \itemsep 0cm \topsep 0cm}}{
        \end{itemize}} 

\newcommand{\dataset}[1]{\textsc{EntailmentBank}}
\newcommand{\mybullet}[1]{$\bullet$ {\bf #1}}

\newcommand{\modelname}[1]{EntailmentWriter}
\newcommand{\taskonename}{no-distractor}
\newcommand{\tasktwoname}{distractor}
\newcommand{\taskthreename}{full-corpus}
\newcommand{\correct}{AllCorrect}
\newcommand{\numenttrees}{1,840}
\newcommand{\entailmentbankurl}{\url{https://allenai.org/data/entailmentbank}}

\newcolumntype{"}{@{\hskip\tabcolsep\vrule width 1pt\hskip\tabcolsep}}
\newcolumntype{'}{@{\hskip\tabcolsep\vrule width 1pt}}
\newcolumntype{!}{@{\hskip 25pt}}
\usepackage{ctable} 

\title{
Explaining Answers with Entailment Trees
}

\author{
Bhavana Dalvi\textsuperscript{*1}, Peter Jansen\textsuperscript{*2}, Oyvind Tafjord\textsuperscript{1},
Zhengnan Xie\textsuperscript{2}, \\
\textbf{Hannah Smith\textsuperscript{2}, Leighanna Pipatanangkura\textsuperscript{2}, Peter Clark\textsuperscript{1}} \\
\textsuperscript{1} Allen Institute for AI, Seattle, WA \\
\textsuperscript{2} University of Arizona, Tucson, AZ \\
\textsuperscript{*} equal contribution \\
\texttt{bhavanad@allenai.org}, \texttt{pajansen@arizona.edu}
}

\eat{
\author{Bhavana Dalvi Mishra, Oyvind Tafjord,\\
\textbf{Peter Clark} \\ 
Allen Institute for AI, Seatle, WA \\
\texttt{\{bhavanad,oyvindt,peterc\}@allenai.org} \\ \And
Peter Jansen, Zhengnan Xie, \\ \textbf{Hannah Smith, Leighanna Pipatanangkura} \\
University of Arizona \\
Tucson, AZ \\
\texttt{pajansen@arizona.edu} \\ }
}

\begin{document}
\maketitle

\begin{abstract}
Our goal, in the context of open-domain textual question-answering (QA), is to
explain answers by showing
the {\it line of reasoning} from what is known to the
answer, rather than simply showing a fragment of textual evidence (a ``rationale'').
If this could be done, new opportunities for understanding and debugging
the system's reasoning become possible.
Our approach is to generate explanations in the form of {\it entailment trees},
namely a tree of multi-premise entailment steps from facts that are known, through intermediate
conclusions, to the hypothesis of interest (namely the question + answer).
To train a model with this skill, we created \dataset{}
\footnote{\dataset{} dataset, annotation tool and evaluation code is available at \entailmentbankurl{}}
, the first dataset to contain
multistep entailment trees. Given a hypothesis (question + answer), we define three increasingly difficult explanation tasks: generate
a valid entailment tree given (a) all relevant sentences (b) all relevant and some irrelevant sentences, or (c) a corpus. 
We show that a strong language model can partially solve these tasks,
in particular when the relevant sentences are included in the input (e.g., $35\%$ of trees for (a) are perfect),
and with indications of generalization to other domains.
This work is significant as it provides a new type of dataset (multistep entailments) and baselines,
offering a new avenue for the community to generate richer, more systematic explanations.

\end{abstract}

\section{Introduction} 

\begin{figure}[t]
\centering
     \includegraphics[width=0.98\columnwidth]{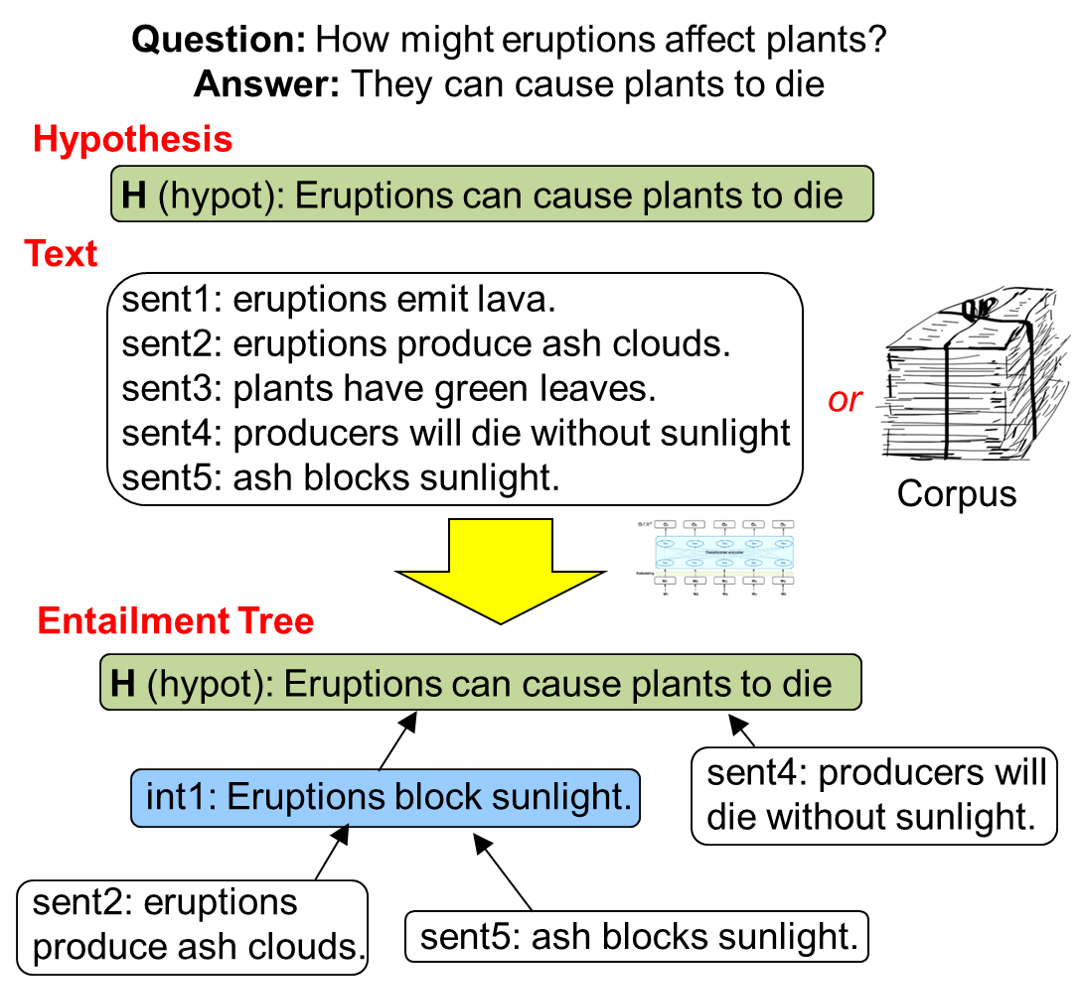}	   
     \caption{Given a hypothesis (green, summarizing a question+answer pair),
       and some partially relevant text (or a corpus), our goal is to
       generate an {\it entailment tree}, including intermediate nodes (blue),
       showing how the hypothesis follows from the text/corpus. \label{example}}
      \vspace{-5mm}
\end{figure}

Explanation remains a formidable challenge in AI. While today's explanation
systems are good at providing a sentence or two of supporting evidence (``rationales'') for an answer \cite{DeYoung2020ERASERAB},
they rarely 
explain the {\it chain of reasoning} from what is known
to the answer, i.e., {\it how} the answer follows, given the evidence --
the goal of this work. Without this, it is hard to fully understand a system's
response and/or pinpoint the source of errors if its conclusions are wrong.
Conversely, if a system could support its answers with a chain of reasoning, new
opportunities arise for interactively teaching the machine by
debugging its mistakes.

Our approach is to generate explanations in the form of multistep {\it entailment trees}, such as shown in Figure~\ref{example},
made up of individual, multi-premise textual entailment (TE) steps \cite{Dagan2013RecognizingTE,Lai2017NaturalLI}.
Although there are many single-step entailment datasets available
\cite{Bentivogli2011TheSP,snli} no dataset of multistep entailments exists,
and so a significant contribution of this paper is the construction
of such a dataset, called \dataset{}. \dataset{} contains
\numenttrees~ multistep entailment trees for accompanying QA pairs, constructed using expert annotators,
and is the first dataset of its kind. We also define three explanation
tasks over this dataset, namely: generate a valid entailment tree for a given QA pair
given (a) all relevant sentences (the leaves of the gold entailment tree), (b) all
relevant and some distractor sentences, or (c) a full corpus. 

%
%
\begin{table*}[t]
\centering
{\small\setlength{\tabcolsep}{2pt}	
\begin{tabular}{l"c|c|c|c"c}  
\specialrule{1pt}{0em}{0em} 
Property$\downarrow$, Dataset$\rightarrow$                      &   WorldTree V2$^1$    &   eQASC$^2$               &   HotpotQA$^3$, R4C$^4$             &   StrategyQA$^5$  &   \textbf{\dataset}  \\\specialrule{0.5pt}{0em}{0em} 
Semantics of Inference          &   \cellcolor{yellow!25} (informal)   &   \cellcolor{green!25} 1-Step Entailment    &   \cellcolor{yellow!25} (informal)  &   \cellcolor{green!25} Deduction   &   \cellcolor{green!25}\textbf{Entailment Tree} \\
Average Facts per Inference     &   \cellcolor{green!25} 5.6                 &   \cellcolor{yellow!25} 2.0                   &   \cellcolor{yellow!25} 2.4             &   \cellcolor{yellow!25} 2.9       &   \cellcolor{green!25}\textbf{ 7.6 }              \\
Average Edges per Inference     &   \cellcolor{green!25} ~~9$^{\ddagger}$ &   \cellcolor{yellow!25} 1                   &   \cellcolor{yellow!25} ~~2$^{\ddagger}$               &   \cellcolor{yellow!25} 2           &   \cellcolor{green!25}\textbf{ 6 }              \\  
Granularity of Inference        &   \cellcolor{green!25} Fine              &   \cellcolor{yellow!25} Coarse              &   \cellcolor{yellow!25} Coarse          &   \cellcolor{yellow!25} Coarse      &   \cellcolor{green!25}\textbf{ Fine }           \\     
Explicit Ordering of Inference  &   \cellcolor{yellow!25} No               &   \cellcolor{yellow!25} No                  &   \cellcolor{yellow!25} No              &   \cellcolor{green!25} Yes         &   \cellcolor{green!25}\textbf{ Yes }             \\
Authoring Method                &   \cellcolor{green!25} Expert             &   \cellcolor{yellow!25} Crowd             &  \cellcolor{yellow!25} Crowd              &\cellcolor{yellow!25} Crowd        &    \cellcolor{green!25}\textbf{ Expert }          \\
\specialrule{1pt}{0em}{0em} 
\multicolumn{6}{r}{
$^1$\cite{xie2020worldtree}
$^2$\cite{Jhamtani2020LearningTE}
$^3$\cite{yang2018hotpotqa}
$^4$\cite{Inoue2020R4CAB}
  $^5$\cite{Geva2021DidAU}}
\end{tabular}}
\vspace{-3mm}
\caption{\small A comparison of \dataset{} with other similar datasets.  In general, \dataset{} contains larger inference problems, at a finer level of granularity than existing datasets, while being the only dataset to include multi-step entailments that make the reasoning steps explicit. $^{\ddagger}$ WT2 and R4C explanations are implied (unannotated) graphs based on overlapping words or entities -- values here are inferred by constructing graphs based on lexical overlap.  \label{dataset-comparison}}
\vspace{-4mm}
\end{table*}

Our focus here is on generating the {\bf derivation} (line of reasoning) showing
how the evidence leads to the answer,
rather than the {\bf pragmatics} of deciding which parts of that to then show
the user. 
This allows us to separate two (typically confounded) explanation requirements,
namely {\it correctness} (of the derivation) from {\it utility}, 
allowing us to evaluate derivations with a more objective measure (correctness).
This also sets the stage for future work on the pragmatics of
what to show users \cite{Miller2019ExplanationIA}. 

Finally, we define and train generative models, called \modelname{}s, for this task, 
adapting earlier techniques 
for generating deductive proofs \cite{proofwriter}. 
We find the models partially solve the dataset, with
indications of generalization to other domains.
\eat{The first model is a 
direct sequence-to-sequence (``all at once'') tree generator. The second model
uses an iterative approach: generate a 1-step entailment from the input text,
add it to the input, and repeat until the target hypothesis is reached or no further
entailments are possible. We find the ``all at once'' approach performs best, but
still at low level, posing a challenge to the community.
}
\noindent
Our contributions are thus:
\vspace{-2mm}
\begin{ite}
\item A formulation of explanation as multistep, multi-premise textual entailment.
\item \dataset{}, the first dataset of multistep entailment trees for QA, to support
entailment-based explanation.  Each tree contains an average of 6.6 nodes and 2.7 entailment steps, 
with the full dataset of \numenttrees~trees including a range of small and large multi-step 
entailment problems. 

\item Baseline results using a state-of-the-art, generative model,
showing that reasonable trees can be generated, in particular
when the necessary raw facts are provided as the model input
(resulting in 35\% of trees with zero errors).
We also present indications that \dataset{}-trained
models can generalize to other domains.
\eat{
\item Baseline results using a state-of-the-art, generative model, showing that 
reasonable trees (42\% with zero errors) can be generated in a restricted setting (all leaf sentences provided),
but that the full task (tree from a corpus) remains hard.  
We include a detailed analysis identifying avenues for future improvement.
}
\end{ite}
This work is significant as it provides a new avenue 
for the community to generate
richer, more systematic explanations.


\section{Related Work}
In the context of QA, there are multiple notions of explanation/justification,
including showing an authoritative, answer-bearing sentence \cite{perez2019finding},
an attention map over a passage \cite{Seo2016BidirectionalAF}, a synthesized phrase
connecting question and answer \cite{Rajani2019ExplainYL}, or the syntactic pattern
used to locate the answer \cite{Ye2020TeachingMC,Hancock2018TrainingCW}.
These methods are primarily designed for answers to ``lookup'' questions, to
explain where/how an answer was found in a corpus.

\eat{
\red{This paragraph could be substantially shortened, given Table~\ref{dataset-comparison} is rich.}
For questions requiring inference, the focus of this paper,
an explanation is sometimes taken as the chain of steps (typically sentences)
leading to an answer. 
Because of the difficulty of generating detailed decomposition steps, existing datasets typically include
one or more simplifications that ultimately limit their utility (see Table~\ref{dataset-comparison} for a comparison). 
For example, HotpotQA's support task goes partway towards this goal
by asking for answer-supporting sentences \cite{yang2018hotpotqa}
or triples (in the derivative R4C dataset \cite{Inoue2020R4CAB}),
but neither dataset describes how these sentences combine to arrive at the answer. 
Similarly, WorldTree V2 \cite{xie2020worldtree} contains supporting sentences but no
information about how they combine. 
More recently, StrategyQA goes further by including derivations for its answers, but using deductive steps and at a coarse granularity \cite{Geva2021DidAU}.
Similarly, eQASC \cite{Jhamtani2020LearningTE} includes reasoning annotations but limited to a single entailment step. In this
work we generalize to tasks requiring {\it multi-step} entailment trees. Table~\ref{dataset-comparison} illustrates these comparisons. 
}

For questions requiring inference, the focus of this paper,
an explanation is sometimes taken as the chain of steps (typically sentences)
leading to an answer. Because crowdsourcing such chains is difficult, existing
datasets typically simplify the task, e.g., collecting answer-supporting sentences
but not how they combine, and/or largely focusing on one-hop (length 2) chains.
Here we generalize to tasks requiring {\it multi-step} entailment trees,
Table~\ref{dataset-comparison} illustrates these comparisons in detail.

\eat{Our dataset also differs from human-authored explanation datasets,
e.g., e-SNLI \cite{camburu2018snli}, and CoS-E \cite{Rajani2019ExplainYL}.
These datasets were primarily designed for (explanation) language generation
(without an associated explanation semantics), while our goal is to assemble 
explanations from an authoritative corpus so that they have credible provenance.}

Our trees are built from {\it multi-premise entailments} (two or more sentences entail a hypothesis), introduced by \citet{Lai2017NaturalLI},
in contrast to the majority of prior datasets where typically a single sentence entails $H$ through (typically) paraphrasing \cite{Bentivogli2011TheSP,BarHaim2014BenchmarkingAS,snli}.
We extend multi-sentence entailment in two ways. First, our trees also show the {\it provenance} of each entailment, namely {\it which} sentences are involved in each entailment
(i.e., going beyond a classification task). Second, ours is the first dataset that chains multiple entailments together into a hypothesis-directed tree,
rather than containing separate, single-step entailments.

Recent work in deductive reasoning has shown that transformers can
generate formal proofs with high reliability, both in a formal
setting \cite{Polu2020GenerativeLM,Wang2020LearningTP} and
with rules expressed in natural language \cite{prover}.
Inspired by this, we apply similar ideas to generating entailment
trees, in particular leveraging
the generative techniques used in the ProofWriter system \cite{proofwriter}
(Section~\ref{models}).




\section{The \dataset{} Dataset}

\begin{figure*}[!th]
	\centering
		\includegraphics[scale=0.925]{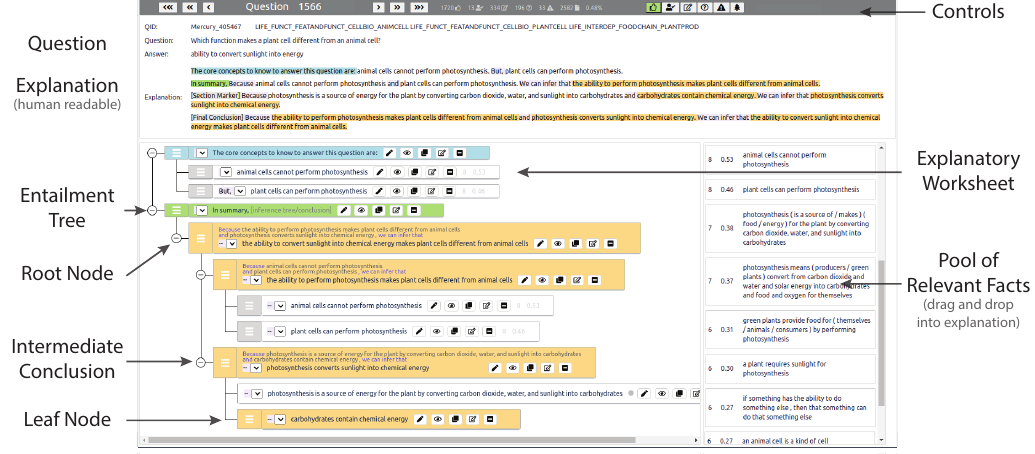}
	\caption{\small \camera{The web-based authoring tool developed to enable authoring entailment trees.}  \textit{(top)} The question and a human-readable version of the semi-structured explanation are provided to the user. \textit{(bottom)} The semi-structured explanation, including the entailment tree, as currently authored by the user.  Nodes (facts) can be dragged-and-dropped to change their ordering.  White nodes represent facts from the corpus, while orange nodes were authored by the user.  \textit{(right)} A shortlist (or pool) of top-ranked relevant facts from the corpus that the user can choose to drag-and-drop into the explanation. }
	\label{fig:main_tool}
\end{figure*}

\dataset{} contains two parts: \numenttrees{} entailment trees, each tree showing {\it how} a
question-answer pair (QA) is entailed from a small number of relevant sentences (e.g., Figure~\ref{example});
and a general corpus $C$, containing those and other sentences of domain-specific and general knowledge
relevant to the QA domain. We use these two parts shortly to define a simpler task (generate the tree given
the leaf sentences, without/with distractors) and a harder task (generate the tree from the corpus).

\dataset{} uses multiple-choice questions (and the correct answer option) from the ARC dataset
of grade-school science questions \cite{clark2018think}, and a corpus of science- and general knowledge
  derived from WorldTree V2 \cite{xie2020worldtree,Jansen2018WorldTreeAC}.
 WorldTree was created for grade-school level science, making it an ideal source for \dataset{}'s corpus.

\subsection{Guidelines}

Three graduate and undergraduate annotators were trained to construct entailment trees for QA pairs, given a small number of
potentially relevant sentences for each QA pair (drawn from WorldTree). Specifically,
they were trained to author trees:
\begin{ite}
 \item where each step is an {\bf entailment} (a conclusion that ``a person would typically infer'' \cite{Dagan2013RecognizingTE}),
i.e., the knowledge expressed in each node reasonably follows from the content of its immediate children.
\item at a {\bf fine-grained granularity}, where each step encodes a single inference, e.g.,
  making a single taxonomic inference, conjoining two facts, or applying a single rule in the corpus.
\item that are {\bf explicit}, with the informal goal of including all the knowledge
  that a young child would need to answer the question.
\item that are {\bf compositional}, where more complex conclusions can be drawn from
  simpler facts.
  \item that are {\bf relevant}, concluding (a declarative version of) the QA pair of interest.
\end{ite}

%
%
\begin{figure}[!t]
	\centering
	\includegraphics[scale=0.45]{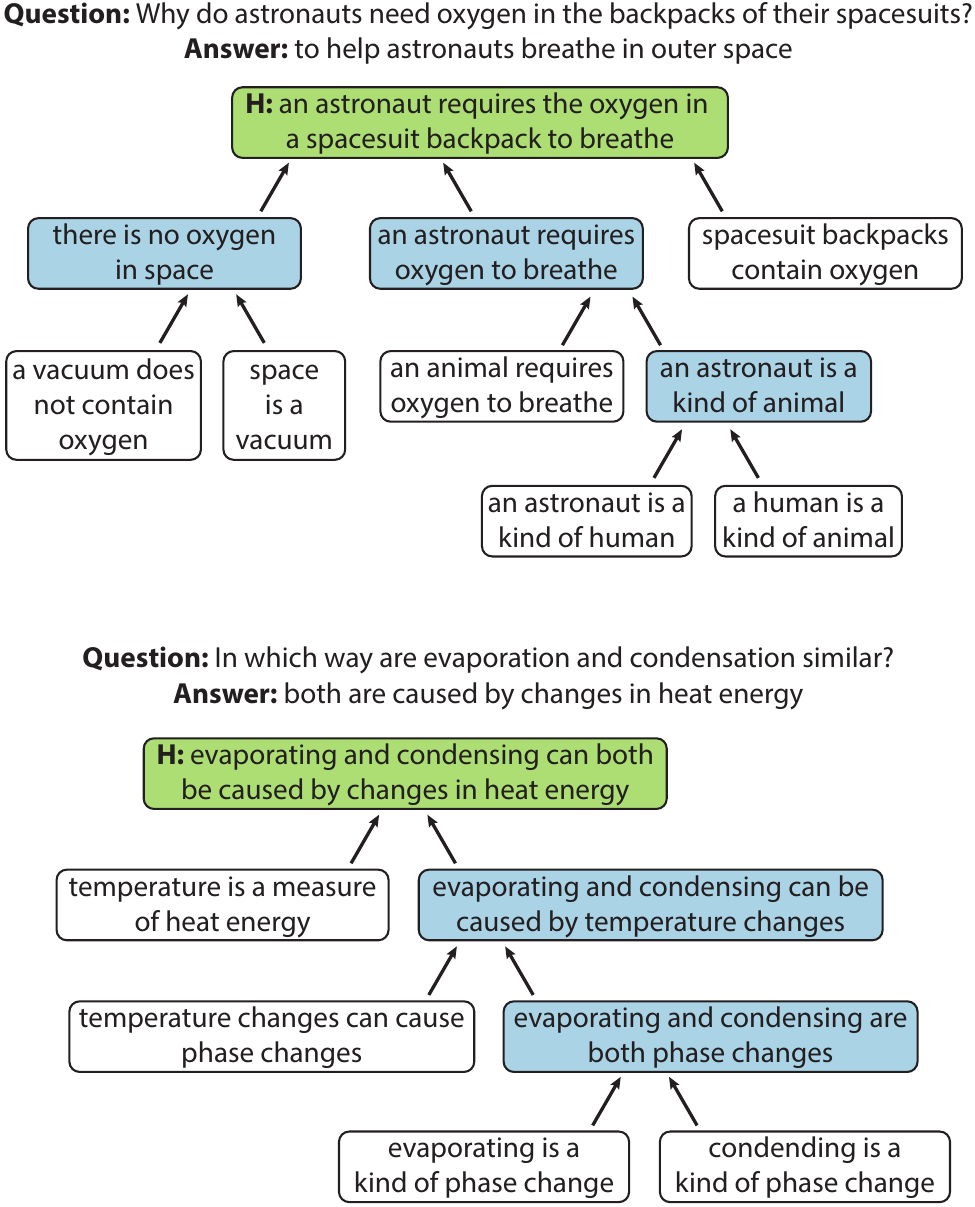}      
	\caption{\small Two example medium-complexity entailment trees, paired with their questions. The root nodes of each tree (hypotheses) are denoted by \textbf{H} (green), and intermediate conclusions are blue. The top tree describes the reasoning to determine why an astronaut requires oxygen in
	spacesuit backpacks, and the bottom to determine the similarity between two concepts (evaporation and condensation).
	\label{fig:tree-examples}}
	\vspace{-14pt}
\end{figure}

\subsection{Tool and Authoring Procedure \label{annotation}}

Constructing detailed entailment trees meeting the above desiderata is challenging.
To make authoring easier, we designed a web-based graphical drag-and-drop authoring tool \footnote{\camera{The \dataset{} authoring tool was implemented as a Javascript browser client and \texttt{npm} back-end, and is released as open source at \url{https://allenai.org/data/entailmentbank}.} }
(\camera{screenshot in Figure \ref{fig:main_tool}})
that allows explanation authors to construct and review explanations quickly.

For each question, the tool presents the user with a pool of top-ranked relevant facts from the corpus\footnote{Details of the retrieval algorithm are in Appendix~\ref{TFRBERT}.} that might be relevant to building an explanation.  To assist in the tree construction process, the user first populates an ``explanatory worksheet'', 
labeling facts that they anticipate will be included in the tree with a small number of specific categories (e.g., \textit{``core facts'', ``grounding facts''}). 
From this worksheet, the user then begins constructing the entailment tree -- typically starting at the bottom-most leaf nodes, authoring intermediate conclusions from them, then progressively working on higher levels of the tree until they author a conclusion that directly answers the question. 

If the user requires a fact not present in the pool of provided facts, 
e.g., a missing science fact or a question-specific statement,
the user can quickly add their own facts and use these in the tree.  
Once completed, the individual entailment steps are then separately reviewed by a different author for quality and suggested edits.  In total, this process takes an average of approximately 20 minutes per question.  Two example trees authored using this process are shown in Figure~\ref{fig:tree-examples}.

\subsection{Overall Dataset}

Due to the large time investment required to generate detailed entailment trees, we author trees for \numenttrees{} randomly selected questions (of the 7,787 in ARC), which include a total of 5,881 discrete entailment steps. Overall, approximately 600 (paid) work hours were used to build the dataset.

%
%
\begin{table}[t]
\centering
{\small
\setlength{\tabcolsep}{3pt}	
\begin{tabular}{lccc|c}  
\specialrule{1pt}{0em}{3pt} 
& Train & Dev & Test & All\\
\specialrule{0.51pt}{3pt}{3pt} 
Questions & 1,313  & 187  &  340 & 1,840\\
Entailment reasoning steps ~~~&  4,175 & 597 & 1,109 & 5,881\\
\specialrule{1pt}{3pt}{0em} 
\end{tabular}
}
\caption{\small Summary statistics for the dataset splits. \label{dataset-stats}}
\end{table}

%
%
\begin{figure}[t]
	\centering
	\includegraphics[scale=0.40]{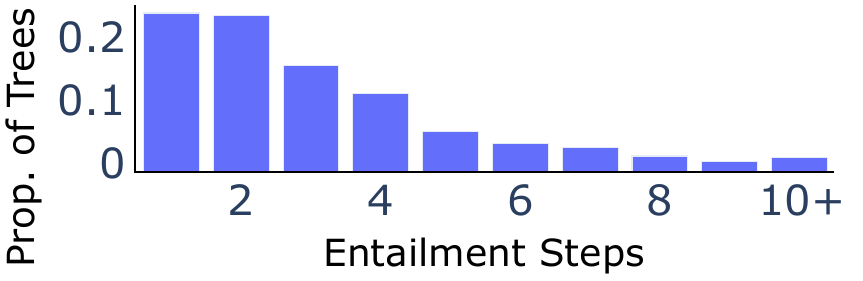}       
	\caption{\small Histogram of entailment steps in the training set.  The average entailment tree contains 7.6 nodes (facts) across 3.2 entailment steps.  \label{fig:tree-statistics}} 
\end{figure}

Summary statistics for the train, development, and test sets are shown in Table~\ref{dataset-stats}.  On average, each entailment tree includes 7.6 nodes across 3.2 entailment steps, where each entailment step typically involves 3 facts (two leaves, that combine to entail a conclusion).  Figure~\ref{fig:tree-statistics} shows a histogram of entailment tree size (measured in terms of number of entailment steps). \dataset{} includes a diverse range of problem sizes, with half (50\%) of entailment trees representing short entailment problems with one or two entailment steps (typically composed of 3-5 nodes), while the remaining 50\% of trees contain 3-17 entailment steps. 




%
%
\newcommand\Tstrut{\rule{0pt}{2.6ex}}         
\newcommand\Bstrut{\rule[-0.9ex]{0pt}{0pt}}   

\definecolor{dthreeblue}{HTML}{1f78b4}
\definecolor{dthreered}{HTML}{e31a1c}
\definecolor{dthreegreen}{HTML}{33a02c}
\definecolor{dthreeorange}{HTML}{ff7f00}


\newcommand{\redB}[1]{\textbf{\textcolor{dthreered}{#1}}}
\newcommand{\blueB}[1]{\textbf{\textcolor{dthreeblue}{#1}}}
\newcommand{\greenB}[1]{\textbf{\textcolor{dthreegreen}{#1}}}
\newcommand{\orangeB}[1]{\textbf{\textcolor{dthreeorange}{#1}}}

\begin{table*}[t]
\centering
{\small\setlength{\tabcolsep}{3pt}	
\begin{tabular}{lccl}  
\textbf{Inference Type}     &   \textbf{Prop.}  &  &   \textbf{Example Entailment} \\
\specialrule{1pt}{0em}{0em} 
Substitution                &   42\%        &   \textcolor{black}{$s_1$}           & when a light wave hits a \redB{reflective object}, the light wave will be reflected       \Tstrut\\
                            &               &   \textcolor{black}{$s_2$}           & a \greenB{mirror} is a kind of \redB{reflective object}       \\
                            &               &   \textcolor{black}{$int$}     & when a light wave hits a \greenB{mirror}, the light wave will be reflected        \\[2pt]            
\hline
Inference from Rule         &   33\%        &   \textcolor{black}{$s_1$}           & if \redB{two species have similar characteristics}, \greenB{they may share a common ancestor}       \Tstrut\Bstrut\\
                            &               &   \textcolor{black}{$s_2$}           & \redB{rhinoceroses and horses have similar characteristics}       \\
                            &               &   \textcolor{black}{$int$}     & \greenB{rhinoceroses and horses might share a common ancestor}       \\[2pt]                                        
\hline
Further Specification or    &   15\%        &   \textcolor{black}{$s_1$}           & an animal requires \redB{warmth} for survival \greenB{as the season changes to winter}       \Tstrut\Bstrut\\
Conjunction                 &               &   \textcolor{black}{$s_2$}           & \redB{thick fur can be used for keeping warm}       \\
                            &               &   \textcolor{black}{$int$}     & \greenB{thick fur can be used for keeping warm as the season changes to winter}       \\[2pt]                                        
\hline
Infer Class from Properties &   4\%         &   \textcolor{black}{$s_1$}           & \greenB{A compound} is \redB{made of two or more elements chemically combined}       \Tstrut\Bstrut\\
                            &               &   \textcolor{black}{$s_2$}           & \blueB{sodium chloride} is \redB{made of two elements chemically combined}       \\
                            &               &   \textcolor{black}{$int$}     & \blueB{sodium chloride} is a kind of \redB{compound}       \\[2pt]            
\hline
Property Inheritance        &   4\%         &   \textcolor{black}{$s_1$}           & \greenB{an animal's shell} is \blueB{usually hard}       \Tstrut\Bstrut\\
                            &               &   \textcolor{black}{$s_2$}           & \blueB{something hard} can be used for \redB{protection}       \\
                            &               &   \textcolor{black}{$int$}     & \greenB{an animal's shell} is \blueB{usually hard} for \redB{protection}       \\[2pt]
\hline
Sequential Inference        &   3\%         &   \textcolor{black}{$s_1$}           & In molecular biology, \orangeB{translation} follows \greenB{transcription}       \Tstrut\Bstrut\\
                            &               &   \textcolor{black}{$s_2$}           & \greenB{transcription} is when genetic information \redB{flows from DNA to RNA}       \\
                            &               &   \textcolor{black}{$s_3$}           & \orangeB{translation} is when genetic information \blueB{flows from RNA to proteins}       \\
                            &               &   \textcolor{black}{$int$}          & In molecular biology, genetic information \redB{flows from DNA to RNA} \blueB{to proteins}  \\[2pt]
\specialrule{1pt}{0em}{0em} 

\end{tabular}}
\caption{\small The prevalence of 6 common reasoning methods required to solve individual entailment tree steps, sampled from 100 random entailment steps in the training corpus.  Discrete entailment steps in \dataset{} require diverse forms of reasoning to solve, from forms of taxonomic or merynomic chaining \textit{(substitution)} to application of domain-specific rules. Here, $s_n$ denotes input sentences, while $int$ denotes entailed conclusions (intermediate nodes in the trees). \label{tab:inference-types}}
\end{table*}

\subsection{Dataset Analysis}

To understand the entailment challenges in \dataset{}, we analyzed
100 randomly sampled entailment steps from trees in the training set.  We identified 6 common high-level categories of inference, shown in Table~\ref{tab:inference-types}.  
\textit{Substitution} types refer to entailments that 
require a model to perform taxonomic, merynomic, or other forms of chaining that substitute one entity for another in one of the input sentences.  \textit{Inference from Rules} entailments 
require the application of a specific rule, specified as one of the input sentences, to the other input sentence.
Our analysis suggests that approximately one-third (33\%) of all entailments require the application of domain-specific rules to complete.  \textit{Further Specification or Conjunction} entailments 
require a model to combine the details of both input facts into a single output fact.  Less frequent types require \textit{inferring an object's class from it's properties}, \textit{inheriting properties of objects}, or determining orders for \textit{sequential reasoning}. 
As a whole, this analysis shows diverse forms of reasoning are required to successfully complete the entailment steps in \dataset{}. 

\eat{
PEC: I think this is already covered in Related Work

\subsection{Comparison with existing explanation datasets?}
average number of reasoning steps, length of questions/answer, explanation granularity etc.
This review can be helpful: \cite{wiegreffe2021teach} \url{https://exnlpdatasets.github.io/structured/}
\todo{PJ: Should we move this comparison section to Related work, like in the QASC paper, as part of the sales pitch?}
}

\section{Task Definitions \label{tasks}}
Because producing correct entailment trees from a corpus is challenging, 
we define three tasks of increasing difficulty 
that simplify the 
problems inherent in the task.
The inputs to all three are a hypothesis $H$, namely a declarative form of a question + answer ($QA$),\footnote{
  For convenience we provide both $H$ and $QA$ as inputs, although in principle $H$ may be generated
  from $QA$ automatically, e.g., using the QA2D model \cite{qa2d}}
and some sentences $S$ expressing (both relevant and irrelevant) knowledge. 
The desired output is a valid entailment tree $T$ where the leaves are sentences selected from $S$,
the intermediate nodes int$_i$ are intermediate conclusions (new sentences, not part of the input), and the
root node (conclusion) is the hypothesis $H$. $T$ is {\it valid} if every node $n_i$ in the tree
is {\it entailed} by its children. 
The 3 tasks vary by the size of $S$, described below.


As an approximation to make automated evaluation feasible, we ensure that $S$ includes all
the leaf sentences $S_{gold}$ that are in the gold entailment tree $T_{gold}$, and treat $T_{gold}$
(+ valid reorderings) as the {\it only} valid entailment tree constructable from
that input. This allows us to check validity by comparing the generated tree with $T_{gold}$.
This approximation is reasonable for tasks 1 and 2 below, because their limited input
makes it unlikely that an alternative valid tree is constructable from the input.
For task 3, though, 
to avoid alternative valid trees being buildable from the input corpus, 
we remove the few sentences similar to $S_{gold}$ from the corpus on a
per-question basis. Although these steps are not fool-proof, they do allow
tree validity to be reasonably approximated by comparing with $T_{gold}$,
a critical requirement for automatic evaluation.

\noindent
The three tasks' inputs are thus as follows:
\begin{des}
\item[{\bf Task 1 (\taskonename):}] Inputs = $H$ + $QA$ + leaf sentences $S_{gold}$
\item[{\bf Task 2 (\tasktwoname):}] Inputs = $H$ + $QA$ + leaf sentences $S_{gold}$ + 15-20 distractor sentences
\item[{\bf Task 3 (\taskthreename):}] Inputs = $H$ + $QA$ + a corpus $C$
 \end{des}
Task 3 represents the full task where $C$ is large. For our experiments, $C$ is the WorldTree corpus plus all additional
science facts created by the annotators (Section~\ref{annotation}).\footnote{
  Some trees also need question-specific scenario facts (e.g., ``A ball rolls down a hill.''), not in $C$ but derivable from $QA$.
  Thus the full Task 3 also requires deriving these. (Our Task 3 baseline does not do this, so has a limitation).}
\eat{
  potentially requiring one or more information retrieval (IR) steps
to construct a valid tree.
In principle, $C$ is simply the WorldTree corpus of general facts. However, in practice
there are two question-specific deviations from this ideal. First, annotators sometimes included non-WorldTree facts in
the gold entailment tree, either to express question-specific scenario facts (e.g., ``A ball rolls down a hill.'')
or additional world knowledge missing in WorldTree. To accommodate this, these extra facts are added to $C$ on 
a per-question basis.
Second, as noted earlier, we remove corpus sentences similar to $S_{gold}$ to discourage alternative valid trees being found, so comparison with $T_{gold}$ is a reasonable evaluation.
}
The desired output in all cases is a valid entailment tree $T$, approximated as being the gold
entailment tree $T_{gold}$ (+ valid reorderings). 

\section{Model \label{models}}


Inspired by the ``All-at-once'' sequence-to-sequence model in the ProofWriter system \cite{proofwriter}, we train three T5-based generative models (one per
task), called \modelname{}s. 

\subsection{Entailment Tree Encoding \label{encoding}}

We encode entailment trees as a linear structure that can be output by a generative model.
To do this, the input sentences $S$ are labeled with identifiers (sent1, sent2, ...),
and the hypothesis $H$ is labeled with the special identifier `hypot' (Figure~\ref{example}).
All nodes in the output tree are then identifiers: sent* for leaf nodes,
int* for internal nodes, and `hypot' for the conclusion (root node). As the int* nodes
denote new sentences (not in the input), we include those sentences in the output 
immediately after their int* identifier is first introduced.


When linearizing the tree, we start from leaf facts and work towards proving the root of the tree (hypot).
We use the symbol ``\&'' to denote ``and'', and ``-$>$'' to denote ``entails''.
Thus the depth 2 entailment tree in Figure~\ref{example} would be encoded as:
\vspace{2mm}
\begin{myquote}
\small{\tt sent2 \& sent5 -$>$ int1: Eruptions block sunlight ; sent4 \& int1 -$>$ hypot}
\end{myquote}
\vspace{1mm}
Note here that the new sentence for intermediate node int1, ``Eruptions block sunlight'', is explicitly part of the to-be-generated output. The task for the models is to output valid entailment trees encoded in this way, given the input.


\subsection{Model Details}
The \modelname{} models are built on top of the text-to-text pretrained T5 transformer \cite{Raffel2020ExploringTL},
where the inputs are as described in Section~\ref{tasks} for Task 1 (\taskonename) and Task 2 (\tasktwoname).
For Task 3 (\taskthreename), the corpus exceeds T5's token limit, so we add a retrieval step of
25 sentences from the corpus $C$ using the hypothesis $H$ as query. 
The output is the predicted entailment tree, encoded as described earlier. 


We fine-tune the models on the training sets using the default hyperparameters 
(including optimizer) 
in the T5 library.\footnote{https://github.com/google-research/text-to-text-transfer-transformer} We use the largest T5-11B model, fine-tuned for 40k steps (batch size 8), 
selecting the checkpoint with highest dev score. 
Additional details about the model can be found in Appendix \ref{appendix:model_selection}.

\section{Experiments}

We train and test three \modelname{}s, one for each task. 
The model inputs are those described earlier for the three tasks, with the exception of Task 3 where a
retrieval step is inserted (the corpus $C$ is too large to be input directly to T5).
For this, we retrieve 25 sentences from $C$ using $QA$ as the query
(using a RoBERTa-trained relevant sentence ranker, details in Appendix~\ref{TFRBERT}),
and input those to the model. The output in all cases is the entailment tree explaining ($H$, the declarative form of) $QA$.

\subsection{Evaluation Metrics \label{metrics}}


We approach evaluating entailment trees as a two step problem.  First, nodes in the predicted tree $T_{pred}$ are aligned with nodes in gold tree $T_{gold}$,
using the sent* labels and Jaccard similarity for intermediate nodes. \camera{Thus, instead of doing exact match against gold tree, we account for semantic-preserving variants (\textbf{Tree Alignment Algorithm} described in Appendix C). 
}

Once aligned, the aligned tree $T_{pred}^{'}$ is scored against gold tree $T_{gold}$ using the metrics below.
The F1/BLEURT metrics score elements of the tree (micro-averaging the results), while ``\correct{}'' 
checks if {\it all} the elements are correct (1=yes, 0=no), i.e., the predicted tree is perfect along
the dimension being considered. Our four metrics are:

\mybullet{Leaf Nodes (F1, \correct{}):}
Does the predicted tree use the correct leaf sentences? We compute an F1 score by comparing leaf sentences $S_{pred}$ to $S_{gold}$. The ``\correct{}'' score is 1 if  
{\it all} nodes are identified correctly (F1=1.0), 0 otherwise.

\mybullet{Steps (F1, \correct{}):}
Are the individual entailment {\it steps} in the tree {\it structurally} correct?
As each intermediate node represents (the conclusion of) a {\it single} step,
the step is considered {\it structurally} correct (score 1) if its input sent*/int* node labels perfectly match the gold, 0 otherwise.
We then measure F1 comparing all steps in the two trees. Then \correct{}=1 if F1=1.0, 0 otherwise.

\eat{
\mybullet{Steps (F1, \correct{}):} Does the tree have the correct entailment step structure?  Here we compute F1 and ``\correct{}'' scores 
in a way similar to leaf nodes by comparing 
the set of entailment steps in the aligned proof $T_{pred}^{'}$ with $T_{gold}$. 
}

\mybullet{Intermediates (F1, \correct{}):} Are the synthesized intermediate nodes  correct?
For comparing gold and generated sentences, we use BLEURT\footnote{
  Using the state-of-the-art BLEURT-Large-512 model. 
  Our analysis based on 300 hand-scored examples suggests its similarity scores
  correlate well with human ratings ({\it correlation=0.67,sensitivity=0.88,specificity=0.80)}}
\cite{sellam2020bleurt}.
We define generation correctness 
as 1 if an aligned pair of $int_{pred}$, $int_{gold}$ gives $BLEURT > 0.28$,\footnote{
The BLEURT threshold was picked using a subset of 300 manually labeled pairs. When
we test this threshold
on the rest of the labeled pairs we get a high (89\%) F1 score, indicating the threshold is reasonable.} 0 otherwise. F1 is computed using the number of aligned, correct intermediates wrt. 
the number of gold/predicted intermediates.
\correct{}=1 if F1=1, otherwise 0.

\eat{
\mybullet{Intermediates (F1, \correct{}):} Are the synthesized intermediate nodes $int_{pred}^{'}$ \correct{}.
For this metric, we use the state-of-the-art BLEURT metric for evaluating generations, using the off-the-shelf 
BLEURT-Large-512 model \cite{sellam2020bleurt}.\footnote{
Our analysis based on 300 hand-labelled examples from the development set suggests
that it correlates well with human ratings ($correlation = 0.67$, $sensitivity=0.88$, $specificity=0.80$).}
We also define generation correctness 
as 1 if an aligned pair of $int_{pred}^{'}$, $int_{gold}$ gives $BLEURT > 0.28$,\footnote{
The BLEURT threshold was picked using a subset of 300 manually labeled pairs. When
we test this threshold
on the rest of the labeled pairs we get a high (89\%) F1 score, indicating the threshold is reasonable.} 0 otherwise. 
Finally, ``Intermediates \correct{}'' per question is  1 if {\it all} the generated intermediates in the tree are correct, 0 otherwise.  
}

\mybullet{Overall Proof (\correct{}):} The overall ``\correct{}'' score for a generated proof is
1 only if all of the leaves, steps, and intermediates are all correct, i.e., the tree
completely matches $T_{gold}$. Otherwise it scores 0. This is a strict metric: any error in
the generated tree will result in a score of 0.

\begin{table*}[t]
\centering
{\small
 \setlength{\tabcolsep}{3pt}	
\begin{tabular}{lp{3pt}ccp{3pt}ccp{3pt}ccp{3pt}c} 
\specialrule{1pt}{0em}{3pt} 
  & \multicolumn{11}{c}{\bf Entailment Tree Scoring} \\
  & ~ & \multicolumn{2}{c}{\bf Leaves} & ~ &\multicolumn{2}{c}{\bf Steps} & ~ & \multicolumn{2}{c}{\bf Intermediates} & ~ & \multicolumn{1}{c}{\textbf{Overall}} \\
  & ~ & {\bf F1}  & {\bf \correct{}}  & ~ & {\bf F1}  & {\bf \correct{}}  & ~ & {\bf F1} & {\bf \correct{}} & ~ & \textbf{\correct{}}\\ \specialrule{0.5pt}{3pt}{3pt} 
{\bf Task 1 (\taskonename)} & ~ & 99.0	& 89.4	& ~ & 51.5	& 38.2	& ~ & 71.2	& 38.5	& ~ & 35.3 \\
{\bf Task 2 (\tasktwoname)} & ~ & 89.1 &	48.8 & ~ &	41.4	& 27.7 & ~ &	66.2 &	31.5 &	~ & 25.6 \\
{\bf Task 3 (\taskthreename)} & ~ &  39.9 &	~3.8 & ~ &	~~7.4 &	~2.9 	& ~ & 35.9	 & 7.1 & ~ &	~~2.9 \\ 
\specialrule{1pt}{3pt}{0em} 
\end{tabular}
}
\caption{\small Baseline scores of the generated entailment trees from \modelname{}, along four different dimensions (test set). 
F1/BLEURT scores measure predicted/gold overlap, while AllCorrect scores 1 when {\it all} the predictions are
correct for a tree, 0 otherwise. 
\camera{Please refer to Appendix Table A2 for scores on the Dev set, and Appendix Table A4 for results using the T5-large.} 
\label{results}}
\vspace{-3mm}
\end{table*}

\subsection{Results}

The results\footnote{Note that the scores are updated from EMNLP 2021 camera ready version.} are shown in Table~\ref{results}. From these, several conclusions can be drawn:

First, in the Task 1 (\taskonename) easiest setting, where only the gold leaves are provided as input,
the {\bf Task1 model performs reasonably well} with over one-third of the trees perfectly matching the gold tree.
From a manual analysis of a random sample of low-scoring trees, we find an additional $\approx$20\% are also
valid but structured differently (thus incorrectly lowering their score), indicating our evaluation metric is an
underestimate. We discuss this in more detail in Section~\ref{tree-analysis}.

Second, Task 2 (\tasktwoname) increases the difficulty by adding distractors
to the input gold sentences until a total of 30 sentences are supplied as input.
Despite this large number of distractors, the model {\bf is good at identifying the relevant facts}
(leaves F1 = 89\%, with nearly half the trees having perfectly selected leaves).
The {\bf overall tree structure in Task2 is (only) a little worse than for Task1} (F1 of steps 41\%, vs. 51\% for Task 1),
despite the substantial additional task complexity.

\camera{Finally, for Task 3, we reuse our Task 2 model (no additional training) but add an IR component to retrieve context from the entire corpus provided for Task 3 (since our model is not able to ingest the entire corpus), using the RoBERTa-based retriever (Appendix~A).
Note that the retrieval is a feature of our baseline system, not of the task specification itself.}

\camera{As shown in Table~\ref{results}, the Task 3 results are lower, indicating that {\bf the full task is difficult}. Although most trees are partially correct in places (e.g., leaf F1 = 39\%), few perfectly match the gold tree. One additional source of error, not present in the earlier Tasks, is that our IR component may not find all the required sentences $S_{gold}$ for the tree. In fact, we find it retrieves 66.1\% of them on average (and also the model input does not include any question-specific scenario facts that may be needed). Thus the lower scores for Task 3 also suggest that the retrieval component is as critical as the tree builder itself (if ingestion of the entire corpus is infeasible); future solutions require either better retrieval or ingestion of the entire corpus. Or, alternatively, a model could generate rather than retrieve some supporting sentences (as illustrated in Figure 4), then use these post-hoc to identify suitable supporting corpus sentences.  }

\subsection{Error Analysis and Future Work}

To understand why invalid trees are sometimes generated, or valid trees mis-scored, we
performed several error analyses that we now describe.

\eat{
To better understand the challenges in generating multistep entailment trees, we performed error analyses both on individual entailment steps, as well as failures associated with generating full trees. 
}
\eat{
We now analyze prediction failures, first at the level of individual entailment steps, then at failures
concerning the overall ``proof'' (tree) structure in the predicted entailment trees.
}

\subsubsection{Individual Entailment Steps}

We first analyze cases where the model is failing at individual entailment reasoning steps.
For this we randomly sampled 100 entailment {\it steps} from imperfect entailment trees (\correct = 0) in the development set.
Manually evaluating these, we found that 30\% were correct entailments (and 13\% were nearly correct),
suggesting {\bf overall invalid trees still contain good steps within them}. 
In cases where the step was invalid, we identify several failure classes and suggest future directions:

\mybullet{Repetition:} \textit{The entailed conclusion simply repeats one of the input sentences (41\%)}, likely because, in many training instances, the intermediate conclusions have high word overlap with input sentences. A \underline{future direction} would be to
modify the loss function to encourage the model to add something novel compared with the input sentences.

\mybullet{Invalid Entailment:} \textit{The entailed conclusion does not follow from input sentences (47\%)}: In these cases, the model is using knowledge unstated in the input for this particular entailment step but present somewhere else in the input context. A \underline{future direction} would be to explore an interative approach, where the model generates one entailment step at a time (a potentially easier entailment task) and then iterates.

\mybullet{Mis-evaluation and Irrelevance:} \textit{The entailed conclusion is correct, but either different from gold or irrelevant to prove the hypothesis (12\%)}. \underline{Future directions} include improving the evaluation metric, and adding a goal-directed term to the loss function to encourage intermediates that are closer to $H$.

\subsubsection{Errors in the Full Entailment Trees \label{tree-analysis}}  
We analyzed an additional 50 imperfect trees on the dev set, and observed the following errors:

\mybullet{Incorrect/missing leaves} ($\approx$50\%):
For example, for the question ``Why do mosquitoes move towards carbon dioxide...? A: It helps mosquitoes find food'', the predicted tree misses using the critical input fact that ``mosquitoes eat animal blood'', hence cannot infer ``animals are a source of food for mosquitoes'', hence cannot infer the importance of moving towards carbon dioxide.
 

\mybullet{Imperfect evaluation} ($\approx$25\%):
We find that a significant number of trees that were scored as invalid are in fact valid, suggesting that our automated
metrics underestimate tree validity. The most common reason was that even with the same input sentences, the tree
can be structured in several valid ways. For example, a gold tree with structure:
\begin{myquote}
  \hspace*{3mm} {\it  sent1 \& sent2 \& sent3 $\rightarrow$ hypot}
\end{myquote}
may be predicted as:
\begin{myquote}
\hspace*{3mm} {\it sent1 \& sent2 $\rightarrow$ int1; int1 \& sent3 $\rightarrow$ hypot}
\end{myquote}
scoring F1=100\% for leaves but F1=0\% for steps, even though valid. (See Appendix~\ref{tree-variation} 
for an instantiated example).
This degree of restructuring is not captured by our metrics.

\hspace*{5mm} To quantify this further, we randomly sampled and rated 50 trees on Task 1 and found human judgements
estimated \textit{Overall AllCorrect} at 58\% (vs. 35.6\% comparing with the gold tree, Table~\ref{results}), suggesting the automated evaluation is
underestimating true task performance by $\approx$20\% in this case.
\underline{Future work} on an improved evaluation metric would help reduce such understimates.

\mybullet{Correct leaves, but invalid steps} ($\approx$20\%): For example, 
for a question asking ``Can a person see someone in a dark room? A: No'', the
model selects the correct leaf sentences but stitches them together in the
wrong order, resulting in invalid intermediate conclusions.
Here, it incorrectly tries to draw an entailment from ``a person is in a dark room'' and ``a person is looking into the dark room'', producing ``the person outside can see the person in the dark room'', an invalid step and one that directly contradicts the target answer.
\underline{Future work} on more reliable entailment, e.g., using an iterative approach and/or adding an entailment validation module, may help address this.

\mybullet{Disconnected trees} ($\approx$5\%): 
We found 2 examples where the generated entailment tree had intermediate conclusions that were not used later towards proving the hypothesis.
\underline{Future work} to avoid this would be to apply structural constraints on the output, enforcing a (single) tree structure.

\mybullet{Correct steps, but incorrect intermediate conclusions}  ($<$5\%): 
For example, for a question with H:``compression waves cause objects to move in the same direction of the wave'', 
the model gets the correct proof structure, but instead of concluding a gold intermediate conclusion ``longitudinal waves are also called compression waves'' it prematurely predicts the final conclusion H for the intermediate (then re-predicts it in the final step).

Finally, we grouped the Task 2 results according to the size (number of steps) in the gold tree.
The results are shown in Appendix Table A3, and demonstrate that the scores drop significantly
as the number of steps in the gold proof increases.  

%

\section{Generality}

To what extent can \dataset{} help for other domains?
Although a full investigation is out of the scope of this paper, we performed two small out-of-domain (OOD) investigations.
First, we used the eQASC dataset \cite{Jhamtani2020LearningTE}.
eQASC provides a QA hypothesis $H$, 10 retrieved sentences $C$, and lists valid {\it single step} entailments from
two sentences in $C$ to $H$ - i.e., one-step (three node) entailment trees.
Running our Task 2 \modelname{} to generate a tree given $C$ and $H$,
we find (comparing with the best-matching gold tree)
a F1 (leaves) of 67\% and an Overall AllCorrect score of 26\% - a positive indication of
transfer OOD. Note that this is without fine-tuning on eQASC, and that eQASC does not 
list all valid entailments, hence good outputs may be missed. 

\begin{figure}[t!]
\small{
\underline{{\bf Input:}} (a QA pair) \\
Q: I would like to go running, but my ankle is injured. \\
\hspace*{4mm} What should I do? \\
A: ice it	\\
\underline{{\bf Output:}}(explanation generated by EntailmentWriter) \\
\fbox{\parbox{1\columnwidth}{
  Icing an injured body part can help to reduce the pain \\
An ankle is a kind of body part \\
$\rightarrow$ Icing an injured ankle can help to reduce the pain }}
\vspace*{2mm} \\
\hspace*{0mm} \underline{{\bf Input:}} (a QA pair) \\
\hspace*{0mm} Q: If I put some cheese in the fridge, will it melt? \\
\hspace*{0mm} A: no, because it will be too cold \\
\hspace*{0mm} \underline{{\bf Output:}}(explanation generated by EntailmentWriter) \\
\fbox{\parbox{1\columnwidth}{
\hspace*{0mm} If something is put in a cold place then it will remain cold \\
\hspace*{0mm} A refrigerator is used for cooling things \\
\hspace*{0mm}  $\rightarrow$ If something is put in a refrigerator then it will remain\\
\hspace*{4mm}                cold. \\
\hspace*{4mm}    Cheese is put in the fridge \\
\hspace*{4mm}     $\rightarrow$ The cheese will remain cold in the refrigerator \\
\hspace*{8mm}        Cold objects do not melt in cold temperatures \\
\hspace*{8mm}        $\rightarrow$ Cheese will not melt in the refrigerator }}
\vspace*{2mm} \\
\hspace*{0mm} \underline{{\bf Input:}} (a QA pair) \\
\hspace*{0mm} Q: I feel sad. What could I do to cheer myself up? \\
\hspace*{0mm} A: listen to music \\
\hspace*{0mm} \underline{{\bf Output:}} (explanation generated by EntailmentWriter) \\
\fbox{\parbox{1\columnwidth}{
  \hspace*{0mm} Music is a kind of entertainment \\
\hspace*{0mm} Entertainment can be used to cheer up a person \\
\hspace*{0mm}   $\rightarrow$  Music can be used to cheer up a person \\
\hspace*{4mm}       If a person is sad, they may want to cheer themselves up \\
\hspace*{4mm}       $\rightarrow$ If a person is sad, they may want to cheer themselves\\
\hspace*{8mm}       up by listening to music }}
}
\caption{\small{Three examples of explanations fully generated by \modelname{}, zero-shot, out-of-domain.
    Here, all sentences (leaves and intermediates) are generated by the model. \label{ood-examples}}}
    \vspace{-5mm}
\end{figure}

We also trained a {\it no-context} version of \modelname{} using \dataset{}, that inputs just a QA pair and
outputs a tree, {\it generating} all the tree sentences (both leaves and intermediates).
We then ran this on Challenge300, an existing, independently authored dataset of 300 test questions covering
multiple domains \cite{macaw}. From a manual evaluation of a random sample of generated trees,
$\approx$35\% were valid, non-vacuous trees. ($\approx$ 25\% of the remainder were
valid but largely repeated the question and answer). Three good examples are shown in Figure~\ref{ood-examples},
again illustrating the potential of \dataset{} for explanation.

\camera{Finally, as an experiment in interactive explanation generation, we re-purposed \dataset{} to train a model to generate an explanation
one step at a time. To do this, we ``shredded'' the entailment trees into individual one-deep trees (where the
intermediate nodes become new hypotheses to prove), and re-trained a model to generate similar one-deep entailment trees.
This model can then be used interactively, generating a one-deep explanation then allowing a user to select
which premise(s) to drill down into, based on what he/she wants to know more about, recursively calling
the model to explain that premise further. Although such generative models (both generating a full tree or
a one-deep tree) can sometimes produce false or nonsensical facts, one could apply 
fact verification techniques, e.g., \cite{Thorne2018FEVERAL,fever2020},
to validate the generated facts, and generate an alternative explanation if validation fails.
These are exciting future directions that we are exploring.}

%

\section{Summary and Conclusion}

Our goal is to enable machines to generate richer, more systematic
explanations. To this end, we have developed
a novel formulation of explanations as {\it multistep entailment trees},
and created \dataset{}, the first large dataset of such trees.

We have also presented baseline results for automatically generating
entailment tree explanations for answers to science questions, trained
on \dataset{}. These initial results suggest that such generation is possible,
in particular when the necessary raw facts are included in the model input.
We have also presented indications that models trained on \dataset{} can generalize to other domains.
This suggests exciting opportunities for future systems that can help
users understand and debug a system's answers, and ultimately
engage in meaningful dialogs that explore the machine's
line of reasoning. \dataset{} contributes to this direction, offering a
new resource for developing richer, more systematic explanations.
\eat{
We have also presented baseline results for automatically generating
entailment tree explanations for answers to science questions, trained
on \dataset{}. These initial results suggest that such generation is possible
in restricted settings (leaf sentences are provided), 
but that the full task (explanation from a corpus) remains challenging.
As new mechanisms are developed to guide and validate tree construction,
we expect the quality of the resulting explanations to improve significantly.
If this were possible, new opportunities for understanding and debugging a
system's reasoning would arise, allowing users to not just receive answers,
but also have meaningful dialogs with the machine about those answers
by exploring its line of reasoning. \dataset{} contributes
to this direction, offering a new resource for developing richer,
more systematic explanations.}
\dataset{} is available at \entailmentbankurl{}.

\section*{Acknowledgements}
\camera{We thank Google for providing the TPUs for conducting experiments. We also thank the Allen Institute of Artificial Intelligence and National Science Foundation award \#1815948 to Peter Jansen for funding this work.}

\bibliography{references}
\bibliographystyle{acl_natbib}

\clearpage

\appendix

\section{Relevant Fact Retrieval Algorithm \label{TFRBERT}}

When authoring an entailment tree for a question, annotators are shown a pool of potentially relevant facts, selected from WorldTree, to help them get started. To identify those facts, we could simply use standard information retrieval with the QA pair as the query. However, for this dataset, we are able to do better than this: First, we train
two ``relevant sentence'' classifiers (using BERT \cite{devlin2019bert} and RoBERTa \cite{roberta} respectively) using additional WorldTree annotations.\footnote{
    WorldTree includes annotations about which WorldTree table rows are relevant to which questions, i.e., which rows
are supporting evidence (``rationales'') for which question. Although these rationales do not identify {\it all} relevant sentences,
they can be used as distant supervision (along with random negative facts drawn from the corpus) to train a ``relevant sentence'' classifier.}
Then, for each question, both models exhaustively score every fact in the corpus, and the top 20 facts from each are retrieved, reranked using Tensorflow-Ranking-BERT \cite{han2020learning}, and presented as a ranked list to the entailment tree annotator based on their final scores.

\section{Evaluation: Tree Alignment Algorithm \label{alignment}}

Predicted entailment trees are evaluated by first aligning them with gold entailment trees, using a variant of the algorithm in \cite{Inoue2020R4CAB}, as follows:
\begin{ite}
\item First, for each intermediate conclusion $int_{pred}$ in $T_{pred}$, and $int_{gold}$ in  $T_{gold}$, 
we gather their ancestor leaf sentences. 
\item Then, we align each intermediate node $int_{pred}$ to the first $int_{gold}$ for which the Jaccard similarity of their respective
ancestor sentences is maximum. For any $int_{pred}$ with zero Jaccard similarity to all gold nodes $int_{gold}$, it is aligned to a dummy gold node with a blank conclusion. 
\end{ite}

\section{Training and Model Selection} \label{appendix:model_selection}
For Task 1 and Task 2, we trained T5 11B models on the training set using default hyperparameters (except the number of steps) following the procedure of Khashabi et al. \shortcite{khashabi2020unifiedqa}.  We used batch size of 8 and a block size of 512 tokens on both input and output side. For both training and evaluation we use v3-8 TPUs from Google cloud computing platform. Each model has 11B parameters and takes 22GB space on disk.

During training, we ran the model for different number of steps (up to 40K steps in the intervals of 4K) and picked the model that gives best Overall AllCorrect score on the Dev set. Thus our hyperparameter search involved 10 models each for Task 1 and Task 2. We picked the models after 16K and 32K steps for Task 1 and Task 2 respectively. Table A2 shows model scores on the development set. 

Each Task required 16 hours of training. 
Inference on 340 test questions takes 12 minutes. A large fraction of this time is spent in saving the model checkpoints to disk or loading the model from disk.

\begin{table*}[t]
\centering
{\small
 \setlength{\tabcolsep}{3pt}	
\begin{tabular}{lp{3pt}ccp{3pt}ccp{3pt}ccp{3pt}c} 
\specialrule{1pt}{0em}{3pt} 
  & \multicolumn{11}{c}{\bf Entailment Tree Scoring} \\
  & ~ & \multicolumn{2}{c}{\bf Leaves} & ~ &\multicolumn{2}{c}{\bf Steps} & ~ & \multicolumn{2}{c}{\bf Intermediates} & ~ & \multicolumn{1}{c}{\textbf{Overall}}  \\
  & ~ & {\bf F1}  & {\bf \correct{}}  & ~ & {\bf F1}  & {\bf \correct{}}  & ~ & {\bf F1}& {\bf \correct{}} & ~ & \textbf{\correct{}}\\ \specialrule{0.5pt}{3pt}{3pt} 
{\bf Task 1 (\taskonename)} & ~ & 99.2	& 90.9	& ~ & 61.8	& 50.3	& ~ & 74.2	& 46.0	& ~ & 42.8 \\
{\bf Task 2 (\tasktwoname)} & ~ & 89.4 &	52.9 & ~ &	46.6	& 35.3 & ~ & 69.1 & 36.9 &	~ & 32.1 \\
{\bf Task 3 (\taskthreename)} & ~ &  42.9 &	~3.7 & ~ &	~~9.0 &	~3.2 	& ~ & 38.4	 & 8.0 & ~ &	~~3.2  \\ 
\specialrule{1pt}{3pt}{0em} 
\end{tabular}
}
\caption*{\small Table A2: Development set results, analogous to test set results for Table~\ref{results}, showing 
baseline scores of the generated entailment trees from \modelname{} along four different dimensions (dev set).}
\label{results_dev}
\end{table*}


%
%
\begin{figure}[t!]
	\centering
	\includegraphics[scale=0.45]{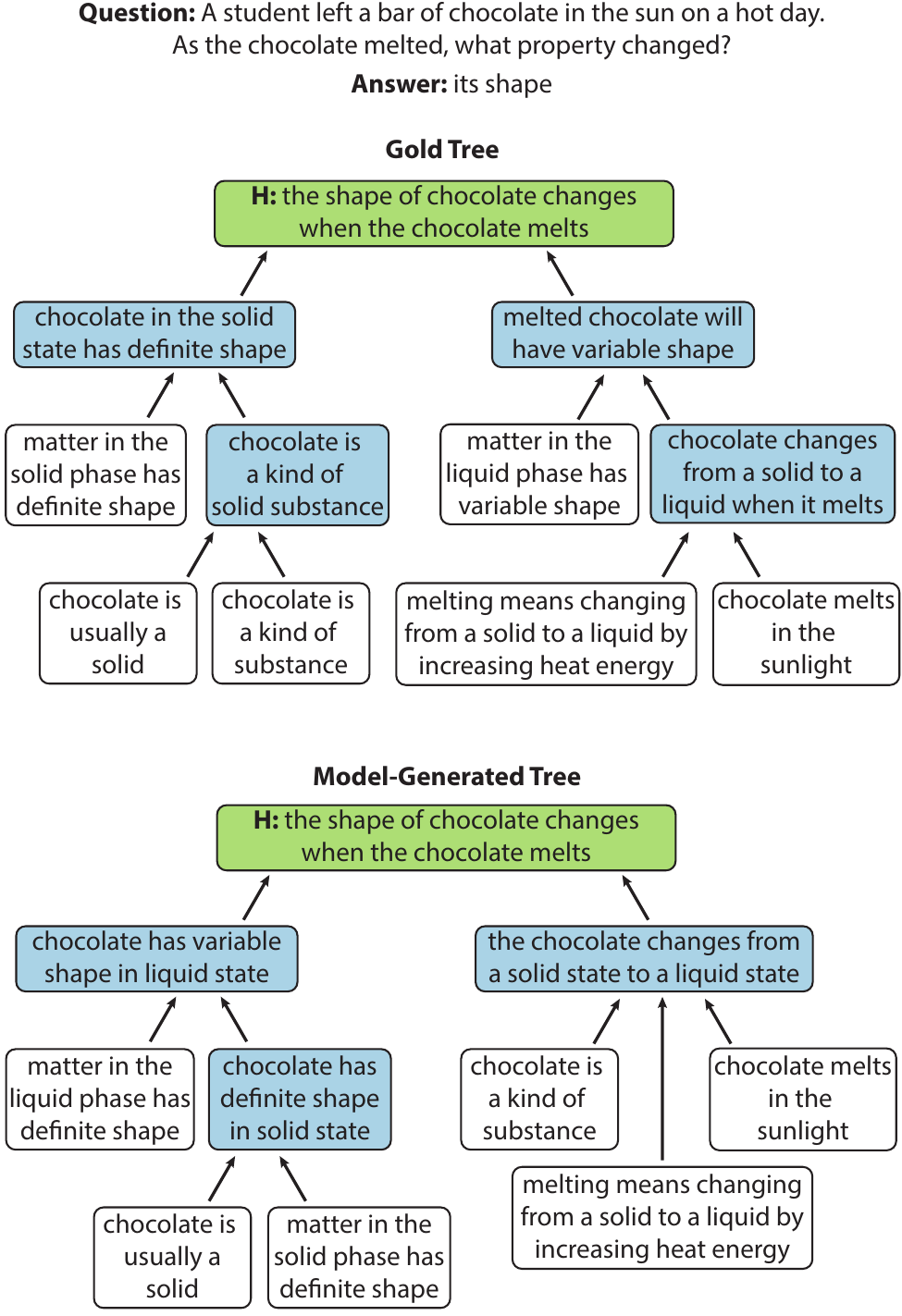}
	\caption*{\small Figure A2: An example question, its gold entailment tree \textit{(top)}, and a model-generated tree \textit{(bottom)} that has different structure and different intermediate conclusions, but is still valid. The root nodes of each tree (hypotheses) are denoted by \textbf{H} (green), and intermediate conclusions are blue. 
    \label{fig:tree-examples-differenttrees}}
	\vspace{-14pt}
\end{figure}

\section{Tree Structure Variation \label{tree-variation}}

\begin{table*}
\centering
{\small
\begin{tabular}{ccccccccc} 
\specialrule{1pt}{0em}{3pt}
\bf{Number of} & \bf{Number of} & \multicolumn{2}{c}{\bf Leaves} & \multicolumn{2}{c}{\bf Steps} & \multicolumn{2}{c}{\bf Intermediates} &  \multicolumn{1}{c}{\textbf{Overall}} \\
{\bf steps} & {\bf questions} & {\bf F1}  & {\bf \correct{}}  & {\bf F1}  & {\bf \correct{}}  & {\bf F1}&  \textbf{\correct{}}&  \textbf{\correct{}}\\ 
\specialrule{0.5pt}{3pt}{3pt} 
1 & 87 & 97.0 & 87.4 & 82.2 & 79.3 & 95.2 & 85.1 & 79.3\\
2 & 84 & 90.5 & 58.3 & 35.0 & 21.4 & 69.5 & 28.6 & 17.9\\
3 & 52 & 87.5 & 32.7 & 25.8 & 5.8 & 59.4 & 7.7 & 0.00\\
4 & 38 & 87.9 & 31.6 & 33.2 & 10.5 & 53.6 & 10.5 & 7.9\\
5 & 28 & 87.3 & 32.1 & 27.9 & 0.00 & 55.4 & 3.6 & 0.00\\
$\geq$6 & 51 & 76.5 & 5.9 & 11.9 & 0.00 & 33.6 & 0.00 & 0.00\\
\specialrule{0.5pt}{3pt}{3pt} 
Any & 340 & 89.0 & 48.8 & 41.4 & 27.6 & 66.2 & 31.5 & 25.6\\
\specialrule{1pt}{3pt}{0em} 
\end{tabular}
}
\caption*{\small Table A3: Results on Task 2 (\tasktwoname) broken down by the number of entailment steps in the gold tree, indicating that scores drop rapidly as trees get larger (more steps).
}  
\label{appendix: results-by-num-steps}
\end{table*}

As described in Section~\ref{tree-analysis}, although our evaluation metric accounts for different node ordering and intermediates wording between the predicted and gold trees, there are still cases where a valid predicted tree differs from the gold tree in a way which (undesirably) hurts its score. For example, a gold tree with the structure:

\begin{table*}[h!]
\centering
{\small
 \setlength{\tabcolsep}{3pt}	
\begin{tabular}{lp{3pt}ccp{3pt}ccp{3pt}ccp{3pt}c} 
\specialrule{1pt}{0em}{3pt} 
  & \multicolumn{11}{c}{\bf Entailment Tree Scoring} \\
  & ~ & \multicolumn{2}{c}{\bf Leaves} & ~ &\multicolumn{2}{c}{\bf Steps} & ~ & \multicolumn{2}{c}{\bf Intermediates} & ~ & \multicolumn{1}{c}{\textbf{Overall}}  \\
  & ~ & {\bf F1}  & {\bf \correct{}}  & ~ & {\bf F1}  & {\bf \correct{}}  & ~ & {\bf F1}& {\bf \correct{}} & ~ & \textbf{\correct{}}\\ \specialrule{0.5pt}{3pt}{3pt} 
{\bf Task 1 (\taskonename)} & ~ &	 98.7 & 86.2	& ~ & 50.5	& 37.7	& ~ & 67.6	& 36.2	& ~ & 33.5 \\
{\bf Task 2 (\tasktwoname)} & ~ & 84.3 & 35.6 & ~ & 35.5 & 22.9 & ~ & 61.8 & 28.5 &	~ & 20.9 \\
{\bf Task 3 (\taskthreename)} & ~ &  35.7 & ~2.9	 & ~ & ~6.1	 & ~	2.4 & ~ & 33.4 & ~7.7  & ~ &	~~2.4  \\ 
\specialrule{1pt}{3pt}{0em}
\end{tabular}
}
\caption*{\small \camera{Table A4: Test set results using \textbf{T5-large} model, analogous to T5-11B results in Table~\ref{results}.}}
\label{results_t5_large_test}
\end{table*}

\begin{myquote}
  \hspace*{3mm} {\it  sent1 \& sent2 \& sent3 $\rightarrow$ hypot}
\end{myquote}
may be predicted as:
\begin{myquote}
\hspace*{3mm} {\it sent1 \& sent2 $\rightarrow$ int1; int1 \& sent3 $\rightarrow$ hypot}
\end{myquote}
scoring F1=100\% for leaves but (undesirably) F1=0\% for steps, even though valid. Figure~A2 shows a more complex example, where both the gold and predicted trees have identical leaf nodes (leaf F1 = 100\%), but different organization. Although both trees are valid, the predicted tree here (undesirably) scores Step F1 = 0\%. Because of cases like this, our predicted scores are an understimate of the true quality of the predictions (by as much as 20\% from a small study, as described in Section~\ref{tree-analysis}).

\section{\camera{Additional Results: T5-large baseline}}

\camera{Here, we trained a T5-large model using default hyperparameters  following the procedure of Khashabi et al. \shortcite{khashabi2020unifiedqa}.  We used batch size of 64 and a block size of 512 tokens on both input and output side. During training, we ran the model for different number of steps (up to 80K steps in the intervals of 8K) and picked the model that gives best Overall AllCorrect score on the Dev set. We picked the models after 48K and 32K steps for Task 1 and Task 2 respectively. Table~A4 shows model scores on the test set. }

\end{document}